# COMPARING METHODS FOR EXTRACTIVE SUMMARISATION OF CALL CENTRE DIALOGUE

Alexandra N. Uma and Dmitry Sityaev

Connex One, 27 Quay Street, Manchester, M3 3GY, United Kingdom

*ABSTRACT*

*This paper provides results of evaluating some text summarisation techniques for the purpose of producing call summaries for contact centre solutions. We specifically focus on extractive summarisation methods, as they do not require any labelled data and are fairly quick and easy to implement for production use. We experimentally compare several such methods by using them to produce summaries of calls, and evaluating these summaries objectively (using ROUGE-L) and subjectively (by aggregating the judgements of several annotators). We found that TopicSum and Lead-N outperform the other summarisation methods, whilst BERTSum received comparatively lower scores in both subjective and objective evaluations. The results demonstrate that even such simple heuristics-based methods like Lead-N can produce meaningful and useful summaries of call centre dialogues.*

## 1. INTRODUCTION

In the last decade, the rate of adoption of AI technology in the contact centre space has been on the increase. The popularity of speech analytics products is not surprising. The technology allows call centre managers to quickly assess the performance of their call centre agents. It also allows analysts to extract business insights from conversations. Businesses can also use information extracted by speech analytics to improve future customer journeys and experiences.

Whilst modern Automatic Speech Recognition (ASR) solutions reach a very high level of accuracy for call transcription, producing accurate and succinct call summaries is still a very challenging task, and the area of conversation summarisation remains an active research field. Call summaries are usually short abstracts summarising the interaction between a customer and an agent. It is not uncommon to capture such important information as a customer's reason for a call, their concerns, agent's handling of the call and final call resolution. Once call conversation transcripts have been obtained, call summarisation can be viewed as a text summarisation challenge.

Various approaches have been proposed and applied towards text summarisation. However, they do fall broadly into two categories: extractive summarisation and abstractive summarisation. Extractive summarisation is based on selecting the most important sentences from the text and presenting them in the summary verbatim (i.e. word for word). Abstractive summarisation is based on creating new paraphrased sentences that summarise the text.

In this paper, we focus on extractive methods for call summarisation. There are several advantages associated with extractive summarisation methods. Firstly, most extractive summarisation methods do not require labelled data (labelling is a very time consuming exercise). Secondly, most extractive summarisation methods do not require any training (with some exceptions). Thirdly, algorithms and models used in extractive summarisation do not normally present challenges for production deployment.

Whilst the task of call summarisation can be addressed through exploring techniques commonly used in text summarisation, it is important to bear in mind that call transcripts exhibit certain characteristics. In particular, call centre conversations normally involve two or more people (e.g. an agent and a customer). Call centre transcripts arising as a result of the application of the ASR technology normally lack punctuation, and punctuation often needs to be restored in order for the summary to be made more comprehensible (Figure 1 illustrates a typical call centre dialogue summarisation pipeline). Additionally, conversations are characterised by such phenomena as hesitations, speech restarts, ill-formed sentences, etc. These aspects need to be additionally addressed when producing call conversation summaries.

In this paper, we evaluate several techniques that can be used to produce call summaries based on call transcripts. We present results from objective and subjective evaluations carried out on our data and point out benefits and limitations of various approaches. The paper is organised as follows. Section 2 briefly surveys previous work on extractive summarisation. Section 3 provides an explanation of how different methods work and the evaluation criteria and tools used. Section 4 provides experimental set up and results for the objective evaluation, whilst Section 5 provides experimental set up and results for the subjective evaluation. Section 6 offers a discussion of the results, stating the limitation of the methods tested. Finally, Section 7 provides a summary of the paper outlining the conclusions, limitations as well as directions for future work.

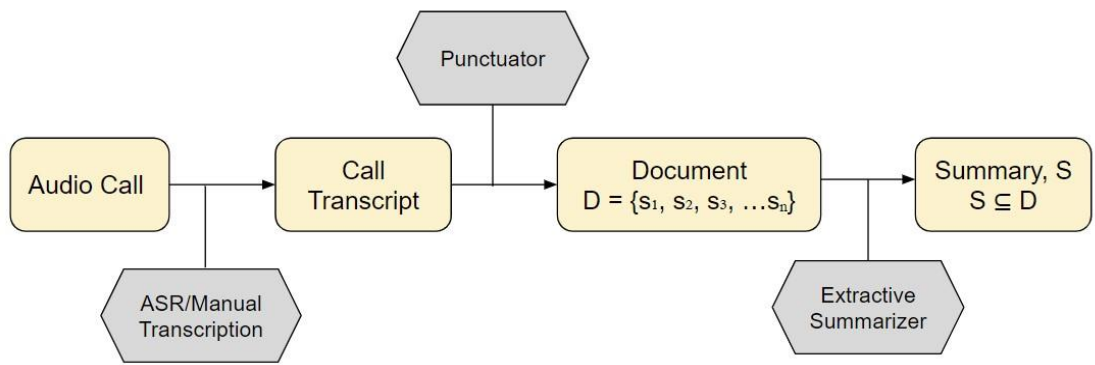

Figure 1. A Typical Call Centre Dialogue Summarization Pipeline

## 2. RELATED WORK

As mentioned above, research on text summarisation is broadly divided into two approaches: extractive summarisation and abstractive summarisation. Below we survey some of the previous work carried out on extractive summarisation.

[1] was one of the first attempts to introduce extractive summarisation for texts. The method is based on extracting important sentences using such features as word frequency and phrase frequency. Common words of high frequencies are ignored in this approach.

[2] and [3] explored the application of HMMs (Hidden Markov Models) for the task of extractive summarisation. The method is based on the model computing the likelihood that a sentence should be included in the summary or not. Only the sentences with maximum posterior probability are selected.

[4] and [5] approached the task of text summarisation using topic models. Latent Semantic Analysis (LSI) technique is used to identify semantically important sentences. Thus, sentences selected for the summary are characterised by minimum redundancy.

[6] attempted to apply fuzzy logic towards the task of text summarisation. 8 most important features are selected and calculated for each sentence first. Fuzzification and inference rules are then applied, with the defuzzification step producing a sentence score. A set of sentences with highest scores is then extracted for the summary.

Neural networks have also been explored for producing summaries. [7] have successfully used RNNs (Recurrent Neural Networks) to carry out extractive summarisation of documents.

More recently, transformer-based models have been considered and applied to text summarisation. [8] used BERT (Bidirectional Encoder Representations from Transformers) model to summarise lecture notes. [9] describes further improvements to the BERT model by way of creating a fine-tuned summariser.

[10] specifically address the task of summarisation of call transcripts for call centres. They propose a novel method which combines call channel separation, topic modelling, sentence selection and punctuation restoration.

A good summary of text summarisation techniques can be found in [11], [12], [13], [14], [15].

## 3. CONCEPTS AND TERMINOLOGY

### 3.1. Lead-N

Lead-N is an extension of the well-used summarisation baseline Lead-3. Lead-N first filters away sentences that contain fewer than 7 non-stop words, then selects the first N number of sentences for the summary. For our purposes, as per user specification, we set the number of sentences for each model at 7. Henceforth, this model would be referred to as Lead-7.

### 3.2. Text Rank

Text rank is a graph-based ranking algorithm proposed by [16] . Text data is split into sentences, and a similarity matrix is created. The similarity matrix is then converted into a graph where sentences are nodes and similarity scores are edges. Top ranked sentences then selected for extractive summary.

### 3.3. KLSum

The KLSum (Kullback-Leibler Sum) algorithm iteratively adds sentences to the summary by selecting at any time step, a sentence that minimises the KL divergence between the candidate summary and the unigram distribution of the original document. An extensive analysis of this model was carried out by [17].

### 3.4. BERTSum

BERTSum is a transformer-based extractive summarisation model proposed by [8]. BERTSum first filters away "too long" or "too short" sentences from the document, then encodes the remaining sentences using the BERT large model, representing each sentence as the averaged embeddings of its tokens. The model then clusters these embedded sentences into N clusters, where N is the number of sentences required for the summary. A summary of the call document is created by taking the centroid sentence for each cluster.

### 3.5. TFIDFSum

TFIDFSum scores each sentence as the sum of the tf-idf scores of the tokens in that sentence. The sentences with the highest scores are then selected for the summary. This method is often used as a baseline summarisation method.

### 3.6. TopicSum

TopicSum, like TFIDFSum, scores each sentence by scoring its tokens. However, rather than using tf-idf scores for tokens, a topic model is applied to the document, and the score of each sentence is the sum of the scores assigned to its tokens by all the topics. We found the Latent Dirichlet Allocation (LDA) topic model [18] with the number of topics set to 15, to perform best for our purposes.

### 3.7. RBMSum

The RBMSum approach proposed by [19] is based on extracting features from each sentence (sentence position, sentence length, etc.). These features are then enhanced using an RBM (Restricted Boltzmann Machine), and sentence scoring is then created using the sum of the enhanced features.

### 3.8. ROGUE-L

ROGUE (Recall-Oriented Understudy for Gisting Evaluation) is a metric often used in objective evaluation of text summaries [20]. The ROUGE-L version of this metric is a function of the longest common subsequence of between the produced summary and the reference summary (also often referred to as the gold summary). Summaries that share a longer sequence with gold summaries tend to have a higher ROUGE-L score. This metric is widely preferred for extractive summarisation evaluation.

### 3.9. MOS

MOS (Mean Opinion Score) is a popular measure for representing the quality of a system or a stimulus. It is a rating which is obtained by averaging scores from a subjective evaluation test.

### 3.10. Label Studio

Label Studio is an open source data labelling tool for labelling video, audio, image, time series, as well as text data[21]. We used this tool to collect annotator judgement in Experiment 2.

## 4. EXPERIMENT 1

In this section, we describe the objective evaluation of the methods discussed in Section 3 using the ROUGE-L metric (see section 3.8). Section 4.1 outlines the details of the experiment such as the dataset used and process of creating reference summaries. Section 4.2 provides the results of the experiments, and we briefly discuss them with the reference to the evaluation metric. A detailed discussion of the results is provided in Section 7.

### 4.1. Method

In this subsection, we discuss the dataset used for this experiment and the procedure for evaluating the models with relation to gold summaries.

#### 4.1.1. Data

For this evaluation experiment, we selected 15 calls with an average duration of 15 minutes at random from the *mobile phone*s domain. For each of these calls we (a) create call documents (b) produce a gold summary from the call document. These steps are detailed below:

**A. Creating Call Documents**

1. Manual Call Annotation. We listened to the calls and manually annotated them using lower case letters only and no punctuation. This is the format typically used by ASR to output call transcriptions.
2. Punctuation Restoration. The next step was to restore punctuation to the call transcripts. We did this using a python package from Hugging Face [22]. This BERT-based punctuator comprises the BERT-base encoder with an additional linear layer. This layer takes the encoded input (the text stream) and for each token predicts whether or not it is followed by a punctuation mark. With punctuation in place, sentence segmentation was carried out using the Spacy Sentencizer [23].

**B. Producing Gold Summaries of Calls**

Reference summaries for each single document version of the call were produced by manually selecting the 7-10 most relevant sentences for each call. We aimed at including the following information in the summary: (1) the reason for the call, (2) pertinent information unique to the call, and (3) the call resolution. We consider these reference summaries as the 'gold standard' for our purposes.

#### 4.1.2. Procedure

The evaluation procedure was as follows:

1. We produced summaries for the 15 documents using the models described in Sections 3.1 - 3.7.
2. We computed the ROUGE-L scores between these summaries and the manually generated gold standard summaries. The results are presented below in Section 4.2.

## 4.2. Results

Table 1 contains the results of evaluating all the models discussed in Sections 3.1 - 3.7 on their ability to produce summaries that closely match the gold summaries using ROUGE-L. As can be seen from the table, although TFIDFSum has a higher recall, Lead-7 outperforms the other models in precision and F1 score.

Table 1. ROUGE-L evaluation of models relative to gold summaries

| Model Name | Precision | Recall | F1 |
|---|---|---|---|
| **Lead-7** | **0.532** | 0.405 | **0.449** |
| TextRank | 0.499 | 0.414 | 0.441 |
| TFIDFSum | 0.460 | **0.428** | 0.429 |
| TopicSum | 0.459 | 0.423 | 0.427 |
| BERTSum | 0.510 | 0.340 | 0.397 |
| KLSum | 0.521 | 0.329 | 0.386 |
| RBMSum | 0.465 | 0.280 | 0.340 |

## 5. EXPERIMENT 2

In this experiment, we evaluated the effectiveness of the summarisation methods using subjective judgements of human annotators. The MOS was used to aggregate these judgements.

### 5.1. Method

This subsection contains a discussion of the data, data preparation and procedure for this subjective evaluation experiment. In this subsection, we discuss the dataset used for this experiment and the procedure for the subjective evaluation of the models.

#### 5.1.1. Data

For this evaluation experiment, we selected 8 calls from 5 domains – *mobile phone*s, *life insurance*, *debt collection, home improvements,* and *solar panel funding*. The average duration of these calls is 11 minutes. The data was processed in the same manner as the data in Experiment 1 (see 4.1.1) and the summaries produced were used for the experiment outlined in 5.1.2.

As a preliminary step, we reduced the number of models from 7 models to 4. The 4 models were selected by examining the summaries produced by the models during Experiment 1. This was done by subjectively ranking the models in order of how well the summaries produced met the gold standard – in other words we asked and answered the question, 'how well did they meet the criteria by which we created gold summaries' (see Section 3.1.1). Based on this, we chose Lead-7, BERTSum, TopicSum and RBMSum. Although TextRank performed competitively with respect to ROUGE-L scoring in Experiment 1, the summaries produced by TextRank for the *life insurance* domain were poor (see Section 6.1). For this reason, it was decided not to include TextRank in subjective evaluation.

We produced summaries of the calls outlined above using the four methods selected.

### 5.1.2. Procedure

Label Studio was used to collect subjective judgements from the annotators about the summaries produced by various models. Our goal was to conduct a 2 hour experiment to ascertain which model's summaries the participants preferred. The participants were drawn from the data science and transcription teams of the company – these participants work on various aspects of call centre data intelligence.

The procedure of this experiment was as follows:

1. We carried out an initial pilot experiment to ensure that the task setup was intuitive. This pilot experiment had 1 participant. This participant was asked to listen to 8 calls (one at a time) and rank each of the four summaries of any given call on a scale of 1-10.
2. We carried out the main experiment with 6 participants. The participants were asked to repeat the steps from the pilot experiment. The names of the models were removed to facilitate an unbiased annotation. It is also important to note that the participants did not have access to the text transcript of the call, or the gold summaries. They had to make their judgement solely on the basis of how well they thought the written summary captured the audio recording of the call.
3. The judgments of the participants were aggregated using the Mean Opinion Score (MOS). The results of this are shown in Table 2 in the section that follows.

### 5.2. Results

Table 2 contains the aggregated results of the subjective judgements of human annotators. The Table places the model name side by side with the mean opinion score for each model (see Section 3.9). As can be seen from the table, TopicSum emerges as the preferred summarisation model, with the highest MOS score of 5.96.

Table 2. Aggregated results of the subjective judgements of human annotators.

| Model Name | Mean Opinion Score |
|---|---|
| **TopicSum** | **5.96** |
| Lead-7 | 5.14 |
| RBMSum | 4.20 |
| BERTSum | 3.66 |

## 6. DISCUSSION

In this section, we discuss the results of the experiments in two parts. Firstly we discuss the suitability of the 7 models evaluated in Section 5. For each method, we analyse the results of Section 6, stating the benefits and limitations of the model. Secondly, we briefly compare subjective and objective evaluation of summaries.

### 6.1. Comparing extractive summarisation methods of call centre dialogue summarisation

Table 1 shows that Lead-7 received the highest F1 score compared to other methods. This is likely because by favouring sentences that occur at the beginning of the call, Lead-7 usually captures the reason for the call, which is an important part of gold summaries. For subjective evaluation, Lead-7 was shown to be the second most preferred method. Taken together, these results show that Lead-7 can be a very competitive baseline for call centre dialogue summarisation. On the negative side, a post-hoc analysis revealed that Lead-7 received lower subjective ratings scores when the sentences in the summary were too long resulting in a "wordy summary". Lead-7 was also revealed to frequently miss the call resolution, a side-effect of selecting sentences from the beginning of the call document.

From Table 1, we see that Lead-7 outperformed TextRank by a small margin when evaluated using ROUGE-L. In fact, a side by side examination of summaries produced by the two models showed that Lead-7 and TextRank produced summaries with the highest degree of token overlap. One possible reason for the competitive ROUGE-L scores for TextRank is that by preferring sentences with the highest similarity to other sentences, TextRank selects sentences with the most diverse words, hence capturing an extensive vocabulary, and Rouge-L being an n-gram based scorer rewards this behaviour. An examination of the summaries produced by TextRank showed them to be highly coherent but with the disadvantage of lacking in topic diversity. Furthermore, for longer calls where several topics are discussed, TextRank often settles on one aspect of the call, ignoring other usually important aspects. For these reasons, TextRank was not included in Experiment 2.

As already mentioned, the heuristic model TFIDFSum works by ranking sentences according to the portion of important tokens they contain: where token importance is computed as token term-frequency inverse document frequency. The competitive ROUGE-L score obtained for this method suggests that by maximising the sum of this token tfidf, TFIDFSum encourages the selection of longer sentences, especially ones that contain weighty words. Also, because high tfidf words are frequently occurring, they are likely to be found across the entire document. Thus, the technique of choosing high-weighted sentences also encourages high coverage summaries. TopicSum, designed to improve on the word importance scoring of TFIDFSum also achieves similar ROUGE-L scores and produces very similar summaries. However, TopicSum produced summaries that contained more call-specific details and for this reason, it was selected for subjective evaluation over TFIDFsum.

Table 1 shows KLSum to have a poor ROUGE-L score – it ranks in the bottom two models according to the metric. As noted in Section 3.2, KLSum aims at producing a summary that best matches the entire document (in our case, the whole call). It does so by selecting a candidate summary with the least divergence from the unigram distribution of the document. This approach, while encouraging diversity, has no notion of topic or word importance and leads to a poor recall of important but succinct topical sentences. Analysis of KLSum summaries revealed they were lacking in coherence, often drifting from topic to topic. This led us to the exclusion of KLSum from Experiment 2.

BERTSum also received a low ROUGE-L score compared to other methods. BERTSum creates N clusters for each document, and selects N sentences closest to each centroid as the document summary. Thus, similarly to KLSum, BERTSum also encourages diversity, however a notion of topicality is enforced by clustering in this case. An examination of BERTSum summaries showed them to also be coherent but lacking in crucial details pertinent to the calls. This lack of detail, possibly a side-effect of the initial filtering process (see Section 3.4) was noted by the annotators in Experiment 2 and is the likely reason for the comparatively low ROUGE-L score for the model.

The last model discussed in the section is RBMSum, which has the worst ROUGE-L score. An examination of the summaries produced by RBMSum shows that RBMSum, like BERTSum, selects shorter sentences, hence producing relatively shorter summaries. That notwithstanding, RBMSum summaries neatly capture the highlights of the calls.

## 6.2. Comparing objective and subjective evaluation

In comparing the objective and subjective evaluation results, we observe two key differences:

1. Although RBMSum received the lowest ROUGE-L score in the objective evaluation, in Experiment 2, the annotators rated RBMSum higher than BERTSum.
2. TopicSum was preferred over Lead-7 by human annotators, even though it scored lower than Lead-7 according to the ROUGE-L metric.

This seems to indicate, in line with findings of the perspectivist view of NLP [24] [25], that the gold standard might be just one opinion of what constitutes a good summary. This suggests that while metrics like ROUGE-L are useful for comparing models, the choice for best summary might be user/purpose dependent.

## 7. CONCLUSIONS AND FUTURE WORK

In this paper, we experimentally compared the use of several extractive summarisation models in building a call centre dialogue summarisation pipeline without depending on gold summaries for model training. To summarise the audio recordings, we convert each call to a document of sentences and apply an extractive summarisation model to extract the most important sentences from this document. We observed that models that take word importance into consideration produce summaries which are most similar to the gold summaries. We also observed that even simple baselines like Lead-7 can produce good summaries of calls. We evaluated the quality of the summaries by aggregating subjective judgements of human annotators. Comparing objective and subjective evaluation of the summaries suggest that both are needed to ascertain the suitability of summarisation models.

We limited the scope of our work in this paper to extractive summarisation techniques as these techniques are often unsupervised and as such neither need labelled training nor present serious challenges for production deployment. Our experiments are also limited in the number of calls annotated, and the expertise of the annotators who, while being well-versed in call centre data transcription and information extraction are not call centre agents or managers. Further research would involve (1) evaluating our results on a larger set of annotated calls of varying lengths with annotation guidelines produced by call centre managers, and (2) extending the scope of our work to include supervised extractive summarization techniques which we will train on dialogue datasets like TweetSum[26] and evaluate on our call centre calls.

### ACKNOWLEDGEMENTS

The authors would like to thank everyone at the Connex One data science team, and the transcription team lead for their participation in this research.

**AUTHORS**

**Alexandra Uma** Alexandra received a PhD in Computer Science from Queen Mary University of London in 2021. She is currently working as a data scientist at Connex One

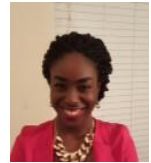

**Dmitry Sityaev** He received MPhil in General Linguistics from the University of Oxford in 1999 and MSc in Data Science from the University of London in 2019. He is currently working as Director of AI at Connex One.

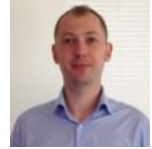